\documentclass[sigconf, nonacm]{acmart}
\usepackage[utf8]{inputenc}
\usepackage{booktabs} 

\begin{document}

\title{Enhancing the QA Model through a Multi-domain Debiasing Framework}

\author{Yuefeng Wang}
\email{yuefeng@utexas.edu}
\affiliation{%
  \institution{The University of Texas at Austin}
  \city{Austin}
  \state{TX}
  \country{USA}
  \vspace{9em}
}

\author{ChangJae Lee}
\email{changjae.lee@utexas.edu}
\affiliation{%
  \institution{The University of Texas at Austin}
  \city{Austin}
  \state{TX}
  \country{USA}
  \vspace{9em}
}

\begin{abstract}
Question-answering (QA) models have advanced significantly in machine reading comprehension but often exhibit biases that hinder their performance, particularly with complex queries in adversarial conditions. This study evaluates the ELECTRA-small model on the Stanford Question Answering Dataset (SQuAD) v1.1 and adversarial datasets AddSent and AddOneSent. By identifying errors related to lexical bias, numerical reasoning, and entity recognition, we develop a multi-domain debiasing framework incorporating knowledge distillation, debiasing techniques, and domain expansion. Our results demonstrate up to 2.6 percentage point improvements in Exact Match (EM) and F1 scores across all test sets, with gains in adversarial contexts. These findings highlight the potential of targeted bias mitigation strategies to enhance the robustness and reliability of natural language understanding systems.
\end{abstract}

\maketitle

\section{Introduction}
Question-answering (QA) systems are widely used in many applications like virtual assistants and customer support tools. These systems answer to the questions by extracting and interpreting information from vast amounts of text. However, despite of significant progress in QA systems, they often inherit biases from the training data. These learned biases can hinder the performance, especially in adversarial scenarios.

\subsection{Problem statement}
QA models are designed to find and pull answers from well-structured texts, but they often struggle when faced with noisy and ambiguous contexts that can mislead the system. QA models are expected to interpret questions and contexts thoughtfully, but they often take shortcuts by learning biases from training dataset. It is required to avoid these biases to make QA systems stronger and more reliable, especially in real-world situations where the data might be unclear or misleading.

\subsection{Objectives}
This study aims to enhance the QA model to be more robust and reliable on complex tasks. To achieve this, error analysis is taken first based on the evaluation results of the baseline model on adversarial sets. After the analysis, a multi-domain debiasing framework is applied to improve the baseline model. Finally, this study quantifies the effectiveness of this framework by comparing the enhanced model’s performance with the baseline.

\section{Analysis}
\subsection{Dataset and model overview}
As a baseline model, the ELECTRA-small model, a transformer-based architecture known for its efficiency and competitive performance in natural language processing tasks \cite{clark2020electra}, is used in this experiment. We trained this model with the Stanford Question Answering Dataset (SQuAD) v1.1, which has over 100,000 questions based on Wikipedia articles \cite{rajpurkar2016squad} for three epochs.
In addition to SQuAD, we utilized two adversarial datasets, AddSent and AddOneSent, which are specifically designed to expose biases and weaknesses by inserting adversarial sentences into contexts \cite{jia2017adversarial}, to evaluate the performance. These datasets challenge models’ ability to generalize and reason beyond superficial patterns.

\subsection{Initial model performance}
The baseline (ELECTRA-small) model achieved strong performance on the original SQuAD dev set, with an Exact Match (EM) score of (78.1 \%) and an F1 score of (85.9\%). However, when evaluated on the adversarial datasets, the scores dropped much, highlighting the model’s susceptibility to adversarial perturbations. On AddSent, the EM score decreased to (52.8\%); on AddOneSent, it was reduced to (62.5\%). These results align with prior findings that challenge datasets can effectively reveal hidden biases in machine learning models \cite{jia2017adversarial}.

\begin{table}[htbp]
\centering
\caption{The baseline (ELECTRA-small) model performance on different datasets}
\label{tab:baseline_perf}
\begin{tabular}{lcc}
\toprule
Dataset & EM score & F1 score \\
\midrule
SQuAD (dev) & 78.1\% & 85.9\% \\
AddSent & 52.8\% & 59.4\% \\
AddOneSent & 62.5\% & 69.1\% \\
\bottomrule
\end{tabular}
\end{table}

\subsection{Error Analysis}
To understand the weaknesses of the baseline model and identify the specific biases affecting its performance on adversarial datasets, we conducted a detailed error analysis. This process involved examining incorrect predictions made by the model on the AddSent and AddOneSent datasets and categorizing them based on predefined bias types. We collected a total of 100 erroneous instances observed in the AddSent and AddOneSent datasets and categorized the errors into three primary types:

\begin{itemize}
    \item Lexical bias
    \item Numerical reasoning errors
    \item Entity recognition and disambiguation errors
\end{itemize}

\subsubsection{Lexical Bias}
Lexical bias occurs when the model relies excessively on word matching between the question and context, leading to incorrect answers, especially in the presence of distractor information that shares similar vocabulary.

\paragraph{Characteristics}
\begin{itemize}
    \item Over-reliance on word matching between questions and context.
    \item Misinterpretation of distractor sentences with lexically similar content.
\end{itemize}

\paragraph{Example}
\begin{itemize}
    \item \textbf{Question:} Where did Super Bowl 50 take place?
    \item \textbf{Context Snippet:} ``\dots Champ Bowl 40 took place in Chicago.''
    \item \textbf{Correct Answer:} Santa Clara, California
    \item \textbf{Predicted Answer:} Chicago
    \item \textbf{Analysis:} The model selects ``Chicago'' due to the phrase ``took place in Chicago,'' demonstrating an over-reliance on lexical overlap despite the context indicating a different event.
\end{itemize}

\paragraph{Impact}
Lexical bias undermines the model's accuracy by prioritizing word similarity over semantic relevance, particularly in adversarial settings where distractors are designed to mislead.

\subsubsection{Numerical Reasoning Errors}
Numerical reasoning errors arise when the model struggles to process, interpret, or reason with numerical data, resulting in incorrect answers involving numbers, dates, or statistics.

\paragraph{Characteristics}
\begin{itemize}
    \item Misinterpretation of numerical values within the context.
    \item Inability to perform basic arithmetic or logical comparisons.
\end{itemize}

\paragraph{Example}
\begin{itemize}
    \item \textbf{Question:} What was the win/loss ratio in 2015 for the Carolina Panthers during their regular season?
    \item \textbf{Context Snippet:} ``\dots The 2020 regular season win/loss ratio for the Michigan Vikings was 656.''
    \item \textbf{Correct Answer:} 15--1
    \item \textbf{Predicted Answer:} 656
    \item \textbf{Analysis:} The model incorrectly selects ``656,'' an irrelevant numerical value from an unrelated sentence, highlighting its inability to reason numerically within the correct temporal context.
\end{itemize}

\paragraph{Impact}
Numerical reasoning errors reduce the model's capacity to answer questions involving numbers, which is critical for handling factual and analytical queries.

\subsubsection{Entity Recognition and Disambiguation Errors}
Entity recognition and disambiguation errors occur when the model misidentifies or confuses entities (such as people, places, and organizations), often due to similar names or contextual proximity within the text.

\paragraph{Characteristics}
\begin{itemize}
    \item Confusion between entities with similar names.
    \item Failure to distinguish between multiple entities mentioned in proximity.
\end{itemize}

\paragraph{Example}
\begin{itemize}
    \item \textbf{Question:} What position does Von Miller play?
    \item \textbf{Context Snippet:} ``\dots Otto Baker plays the position of hamster.''
    \item \textbf{Correct Answer:} Linebacker
    \item \textbf{Predicted Answer:} hamster
    \item \textbf{Analysis:} The model confuses the relevant entity (``Von Miller'') with an unrelated mention (``Otto Baker''), leading to an incorrect and nonsensical answer.
\end{itemize}

\paragraph{Impact}
Entity recognition errors impair the model's ability to associate information with the correct entity, leading to factual inaccuracies.

\subsection{Error Distribution}
Analysis of 100 errors sampled on the AddSent and AddOneSent datasets revealed clear patterns in three categories. Most errors (61\%) come from lexical bias, followed by numerical reasoning errors (25\%) and entity recognition errors (14\%).

\begin{table}[h]
\centering
\caption{Error categories and distribution}
\label{tab:error_dist}
\resizebox{\columnwidth}{!}{%
\begin{tabular}{lcc}
\toprule
Error Type & Number of Errors & Percentage \\
\midrule
Lexical bias & 61 & 61\% \\
Numerical reasoning errors & 25 & 25\% \\
Entity recognition and disambiguation errors & 14 & 14\% \\
\midrule
Total & 100 & 100\% \\
\bottomrule
\end{tabular}%
}
\end{table}

The high rate of lexical bias suggests that the model often depends on matching words between the questions and context. This reliance leads to wrong answers when similar-sounding distractor information is present. This finding highlights a major weakness in the model's understanding, particularly in separating relevant information from similar but unimportant content.

\textit{Note: While this analysis provides valuable insights into common error patterns, the sample size of 100 errors represents only a small subset of the dataset. As such, the percentages should not be interpreted as definitive for the entire dataset. Instead, they offer an indicative overview of the biases affecting the model's performance.}

\section{Methodology}

\subsection{Framework}
To improve and generalize the QA model, the multi-domain debiasing framework \cite{wu2020improving} is adopted in this experiment. This framework focuses on expanding the dataset domain to learn more general QA tasks and down-weighting biased samples to avoid learning biases. Three main strategies are used as follows: 1) knowledge distillation, 2) debiasing, and 3) expanding domains.

\subsection{Knowledge Distillation}
Knowledge distillation is to train a student model by learning from an ensemble of teacher models in multiple domains or a teacher model in a single domain. A student model learns from the soft labels predicted by teacher models, the probability distribution of start logits and end logits, rather than the one-hot labels from the training set. This approach distills the teacher models' knowledge into a student model and can capture richer structural information even with a single domain by using soft labels \cite{mobahi2020self, hinton2015distilling}.

\subsection{Debiasing}
As we examined error patterns of the baseline model in the error analysis section, a model can learn biases from training data. This result leads a model to be less accurate in answering out-domain questions. A down-weighting method that smoothes the signals from biased examples is applied to minimize their effects.

In order to avoid learning bias patterns, biased examples need to be recognized first. Bias models that only exploits bias patterns to answer the question are used to detect biased examples \cite{clark2019dont, mahabadi2020end, utama2020mind, he2019unlearn} here. Bias weights of each example, representing how much the question can be answered correctly depending on its biases, are obtained from bias models. Bias weights are the predicted probability of the start and end tokens, respectively, when the prediction from a bias model is correct; otherwise, it is zero. We used pre-calculated bias weights \cite{wu2020improving} for each example in this experiment.

The confidence regularization method \cite{utama2020mind} is used as a down-weighting method for debiasing. This method smoothes the predicted probability distribution from the teacher model based on bias weights.

Table \ref{tab:debiasing_results} shows the EM and F1 scores from the debiased model compared to the baseline model from Section 2. The debiased model is based on pre-trained ELECTRA-small as well, and the baseline model is used as a teacher model. The debiased model has been trained for three epochs. The results have improved modestly, around 1 percentage point for EM and F1 scores, for all datasets, including the SQuAD dev set and adversarial sets.

\begin{table}[h]
\centering
\caption{Comparison of EM and F1 scores between baseline model and debiased model on SQuAD dev set and Adversarial sets}
\label{tab:debiasing_results}
\resizebox{\columnwidth}{!}{%
\begin{tabular}{lcccc}
\toprule
 & \multicolumn{2}{c}{EM Score} & \multicolumn{2}{c}{F1 Score} \\
\cmidrule(lr){2-3} \cmidrule(lr){4-5}
Dataset & Baseline & Debiased & Baseline & Debiased \\
\midrule
SQuAD (dev) & 78.1\% & 79.0\% (+0.9) & 85.9\% & 86.5\% (+0.6) \\
AddSent & 52.8\% & 53.5\% (+0.7) & 59.4\% & 60.4\% (+1.0) \\
AddOneSent & 62.5\% & 63.1\% (+0.6) & 69.1\% & 70.4\% (+1.3) \\
\bottomrule
\end{tabular}%
}
\end{table}

From the result, we can see that debiasing can help improve prediction. However, since SQuAD is relatively clean and well-structured with relevant context, it is hard to improve predictions on adversarial sets with noisier and irrelevant contexts only using the debiasing method.

\subsection{Expanding Domains}
With an ensemble of teacher models from several domains, building a more robust and generalized QA model is possible. Distilled knowledge from multiple domains can help a model predict under adversarial settings more accurately. In addition to SQuAD, we expand our domain with the HotpotQA \cite{yang2018hotpotqa}, which has more complex contexts and multi-hop reasoning tasks.

Table \ref{tab:two_domain_results} shows the EM and F1 scores from the debiased model under two-domain settings, SQuAD and HotpotQA, compared to the baseline model from Section 2. The teacher model for HotpotQA is also based on the pre-trained ELECTRA-small, and the student model has been trained for three epochs. The results have improved more than those from a single domain, and EM and F1 scores for adversarial sets have increased by around 2 percentage points compared to the baseline.

\begin{table}[h]
\centering
\caption{Comparison of EM and F1 scores between Baseline model and Debiased model under two domain settings (SQuAD + HotpotQA)}
\label{tab:two_domain_results}
\resizebox{\columnwidth}{!}{%
\begin{tabular}{lcccc}
\toprule
 & \multicolumn{2}{c}{EM Score} & \multicolumn{2}{c}{F1 Score} \\
\cmidrule(lr){2-3} \cmidrule(lr){4-5}
Dataset & Baseline & Two Domains & Baseline & Two Domains \\
\midrule
SQuAD (dev) & 78.1\% & 78.9\% (+0.8) & 85.9\% & 86.8\% (+0.9) \\
AddSent & 52.8\% & 55.0\% (+2.2) & 59.4\% & 61.3\% (+1.9) \\
AddOneSent & 62.5\% & 64.1\% (+1.6) & 69.1\% & 70.9\% (+1.8) \\
\bottomrule
\end{tabular}%
}
\end{table}

We expand our domain to more datasets under MRQA frameworks \cite{fisch2019mrqa} to see how it can generalize the QA model. TriviaQA \cite{joshi2017triviaqa}, NewsQA \cite{trischler2017newsqa}, and Natural Questions (NQ) \cite{kwiatkowski2019natural} are added to the dataset domain. Pre-calculated bias weights of each example from Wu \cite{wu2020improving} are used for debiasing. Table \ref{tab:bias_ratio} shows the ratio of biased examples in each domain, which is provided by Wu \cite{wu2020improving} using bias models.

\begin{table}[h]
\centering
\caption{The ratio of biased examples in each domain}
\label{tab:bias_ratio}
\resizebox{\columnwidth}{!}{%
\begin{tabular}{lccccc}
\toprule
Dataset & SQuAD & HotpotQA & TriviaQA & NewsQA & NQ \\
\midrule
\% of Biased samples & 61.9\% & 74.5\% & 58.1\% & 31.8\% & 64.8\% \\
\bottomrule
\end{tabular}%
}
\end{table}

Table \ref{tab:five_domain_results} shows the results of a debiased model under five-domain settings compared to the baseline model. All teacher models are based on the ELECTRA-small, and the student model has been trained for two epochs.

\begin{table}[h]
\centering
\caption{Comparison of EM and F1 scores between baseline model and debiased model under five domain settings}
\label{tab:five_domain_results}
\resizebox{\columnwidth}{!}{%
\begin{tabular}{lcccc}
\toprule
 & \multicolumn{2}{c}{EM Score} & \multicolumn{2}{c}{F1 Score} \\
\cmidrule(lr){2-3} \cmidrule(lr){4-5}
Dataset & Baseline & Five Domains & Baseline & Five Domains \\
\midrule
SQuAD (dev) & 78.1\% & 79.9\% (+1.8) & 85.9\% & 87.4\% (+1.5) \\
AddSent & 52.8\% & 55.0\% (+2.2) & 59.4\% & 62.0\% (+2.6) \\
AddOneSent & 62.5\% & 64.4\% (+1.9) & 69.1\% & 71.5\% (+2.4) \\
\bottomrule
\end{tabular}%
}
\end{table}

The prediction results have improved over two domain settings for all test sets, especially in F1 scores. The improvement in adversarial sets might appear moderate since our approach does not specifically target learning from adversarial settings. Instead, our method focuses on enhancing the model's generalization ability across various QA settings, while adversarial training can degrade the performance on general settings \cite{kaushik2021efficacy, park2024adversarial}.

\section{Results}
As shown in the last section, the debiased models consistently outperform the baseline model across all test sets. The improvements in the adversarial datasets, AddSent and AddOneSent, indicate that the bias mitigation strategies effectively enhance the model's robustness against adversarial perturbations. Additionally, the experiments on expanding domains demonstrated that adding more domains help generalize and build a more reliable QA model.

\subsection{Error Reduction Analysis}
To evaluate the impact of the debiasing strategies on specific biases, we revisited the sample of 100 errors analyzed in Section 2.4. After applying the mitigation strategies, 7 errors were corrected: 4 lexical bias errors and 3 numerical reasoning errors, while no improvements were observed in entity recognition errors.

\begin{table}[h]
\centering
\caption{Number of errors corrected across different bias categories after applying mitigation strategies}
\label{tab:error_reduction}
\resizebox{\columnwidth}{!}{%
\begin{tabular}{lccc}
\toprule
Error type & Baseline errors & Errors corrected & Reduction (\%) \\
\midrule
Lexical bias & 61 & 4 & 6.6\% \\
Numerical reasoning errors & 25 & 3 & 12.0\% \\
Entity recognition errors & 14 & 0 & 0.0\% \\
\midrule
Total & 100 & 7 & 7.0\% \\
\bottomrule
\end{tabular}%
}
\end{table}

Additionally, this analysis is based on a small sample of 100 errors, representing only a fraction of the dataset. Therefore, the results should be viewed as indicative rather than definitive.

\subsection{Examples of Corrected Errors}

\paragraph{Lexical Bias}
\begin{itemize}
    \item \textbf{Question:} ``Who won Super Bowl 50?''
    \item \textbf{Context Snippet:} ``\dots Stark Industries won Champ Bowl 40.''
    \item \textbf{Debiased Prediction:} ``Denver Broncos''
    \item \textbf{Correct Answer:} ``Denver Broncos''
    \item \textbf{Analysis:} The baseline model was misled by the adversarial sentence due to lexical overlap. The debiased model correctly identifies ``Denver Broncos'' as the winner of Super Bowl 50.
\end{itemize}

\paragraph{Numerical Reasoning}
\begin{itemize}
    \item \textbf{Question:} ``What was Ronnie Hillman's average yards per carry in 2015?''
    \item \textbf{Context Snippet:} ``\dots and a 4.7 yards per carry average. Boyd Holman's average yards per carry in 2020 was 9.7.''
    \item \textbf{Debiased Prediction:} ``4.7''
    \item \textbf{Correct Answer:} ``4.7''
    \item \textbf{Analysis:} The debiased model accurately selects the correct numerical value despite the presence of a distractor.
\end{itemize}

\section{Discussion}
The application of the multi-domain debiasing framework has led to measurable improvements in the ELECTRA-small model's performance. While the overall error reduction is modest, the enhancements in EM and F1 scores demonstrate the effectiveness of our mitigation strategies in strengthening the model's comprehension and robustness \cite{hinton2015distilling}.

However, despite the improvements, several limitations persist in this study:

\begin{enumerate}
    \item \textbf{Small sample size in Error Analysis:} The error analysis relied on a sample of 100 errors, which provides indicative insights but may not fully represent trends across the entire test set.
    \item \textbf{Entity recognition gaps:} No improvements were observed for entity recognition errors, indicating the need for advanced Named Entity Recognition (NER) systems or coreference resolution techniques to address these challenges \cite{lample2016neural, devlin2019bert}.
    \item \textbf{Limited gains in adversarial datasets:} While F1 and EM scores improved, gains in adversarial contexts (e.g., AddSent, AddOneSent) were modest, suggesting a need for complementary approaches like adversarial training.
    \item \textbf{High computational costs:} Expanding domains and using ensemble models increases computational demands, which may limit scalability in resource-constrained environments.
\end{enumerate}

\section{Conclusion}
This study systematically identified and addressed key biases affecting the ELECTRA-small model's performance on machine reading comprehension tasks. By categorizing errors into lexical bias, numerical reasoning errors, and entity recognition and disambiguation errors, we implemented a multi-domain debiasing framework comprising knowledge distillation \cite{hinton2015distilling}, debiasing techniques, and domain expansion. These targeted strategies collectively enhanced the model's accuracy and robustness across both general and adversarial datasets \cite{jia2017adversarial}.

Although the framework successfully reduced some biases, challenges remain, particularly in addressing entity recognition errors and achieving larger gains in adversarial contexts. Future work should focus on integrating advanced NER techniques \cite{lample2016neural, devlin2019bert}, expanding error analysis to larger datasets, and optimizing computational efficiency.

In conclusion, this research highlights the importance of targeted debiasing strategies for building more robust and reliable QA systems capable of handling complex and adversarial queries.

\bibliographystyle{ACM-Reference-Format}
\bibliography{ref}

\end{document}